\documentclass[10pt,twocolumn,letterpaper]{article}

\usepackage{cvpr}
\usepackage{times}
\usepackage{epsfig}
\usepackage{graphicx}
\usepackage{amsmath}
\usepackage{amssymb}
\usepackage{algorithm}
\usepackage{listings}
\usepackage{tcolorbox}
\usepackage{tabularx}
\usepackage{algpseudocode}

\usepackage[pagebackref=true,breaklinks=true,letterpaper=true,colorlinks,bookmarks=false]{hyperref}

\cvprfinalcopy % *** Uncomment this line for the final submission

\def\cvprPaperID{****} % *** Enter the CVPR Paper ID here

\ifcvprfinal\fi
\begin{document}

%%%%%%%%% TITLE
\title{Deep Learning Based Dense Retrieval: A Comparative Study}

\author{
  Ming Zhong\\
  {\tt\small mzhong45@gatech.edu}
  \and
  Zhizhi Wu\\
  {\tt\small zwu435@gatech.edu}
  \and
  Nanako Honda\\
  {\tt\small nhonda3@gatech.edu}
  \and
  \\
  \multicolumn{1}{c}{\textit{Georgia Institute of Technology}}\\
  \multicolumn{1}{c}{North Avenue, Atlanta, GA 30332}
}

\maketitle
%\thispagestyle{empty}

%%%%%%%%% ABSTRACT
\begin{abstract}
   Dense retrievers have achieved state-of-the-art performance in various information retrieval tasks, but their robustness against tokenizer poisoning remains underexplored. In this work, we assess the vulnerability of dense retrieval systems to poisoned tokenizers by evaluating models such as BERT, Dense Passage Retrieval (DPR), Contriever, SimCSE, and ANCE. We find that supervised models like BERT and DPR experience significant performance degradation when tokenizers are compromised, while unsupervised models like ANCE show greater resilience. Our experiments reveal that even small perturbations can severely impact retrieval accuracy, highlighting the need for robust defenses in critical applications. 
\end{abstract}

%%%%%%%%% BODY TEXT
\section{Introduction/Background/Motivation}

For a long time, the Information Retrieval (IR) field has been dominated by sparse retrieval systems, which match texts based on lexical patterns (e.g., BM25) \cite{Okapi}\cite{bm25}. However, recent advancements in deep learning, coupled with Retrieval Augmented Generation (RAG), have revolutionized text generation. RAG allows models to access curated external knowledge databases, ensuring that the generated text is grounded in factual information. This has led to the development of dense retrievers, which match texts based on semantic meanings. Dense retrievers, which use deep learning combined with RAG, can automatically learn query and document representations from labeled data, giving them exceptional capabilities for capturing and representing text semantics for relevance matching \cite{Dense-text-retrieval-based-on}.

\subsection{Objectives}
Our objective is to investigate how dense retrievers perform when their tokenizers are tampered with or "poisoned". A poisoned tokenizer introduces malicious or biased behavior into the model, potentially leading to incorrect or misleading tokenization of input text, and thus affecting the model's ability to understand and process text accurately. We aim to assess the impact of poisoned tokenizers on dense retriever models and explore potential mitigation strategies.

\subsection{Current Practice}
Currently, dense retrievers rely heavily on robust tokenizers to break down input text into numerical tokens that models can process. This tokenization is crucial for the model’s understanding and subsequent retrieval performance \cite{mielke2021wordscharactersbriefhistory}. However, if the tokenizer is compromised, the model’s performance can degrade significantly. Current practices do not adequately address the robustness of tokenizers against such attacks, leaving a gap in ensuring the reliability and security of dense retrievers.

\subsection{Significance}
Understanding the impact of poisoned tokenizers is crucial for improving the security and reliability of dense retrieval systems. If successful, our research will highlight the vulnerabilities of current models and suggest ways to enhance their robustness. This will be particularly beneficial in applications where the integrity of information retrieval is critical, such as in legal, medical, and financial domains.

\subsection{Data}
We use the following datasets for our research:

\begin{itemize}
    \item \textbf{Quora Dataset} \cite{Quora}: This dataset contains over 400,000 question pairs, each labeled to indicate whether the questions are paraphrases of each other.
    \item \textbf{Financial Opinion Mining and Question Answering (FiQA-2018)} \cite{FiQA-2018}: Crafted for the FiQA challenge in 2018, it includes opinion-based questions and answers tailored to the financial domain.
    \item \textbf{HotpotQA} \cite{HotpotQA}: A question answering dataset that features multi-hop questions, requiring systems to establish logical connections among diverse pieces of information to provide accurate answers.
\end{itemize}

These datasets were chosen for their diversity and relevance in evaluating the robustness and performance of dense retrievers under the influence of poisoned tokenizers.

\section{Approach}

% (10 points) What did you do exactly? How did you solve the problem? Why did you think it would be successful? Is anything new in your approach? 

% (5 points) What problems did you anticipate? What problems did you encounter? Did the very first thing you tried work? 

We first explain the evaluation metrics we use in details. Then, we describe target models and the evaluation design to assess the effectiveness of supervised and unsupervised (especially contrastive learning) dense retrieval models.

\subsection{Evaluation metrics}
\paragraph{Cosine Similarity} Cosine similarity is used to measure the similarity between two sentence embeddings \cite{vaswani2023attentionneed}\cite{devlin2019bertpretrainingdeepbidirectional}\cite{reimers-2019-sentence-bert}\cite{karpukhin-etal-2020-dense}. Given two sentence embeddings $\mathbf{u}$ and $\mathbf{v}$, cosine similarity is defined as:

\begin{equation}
\text{similarity}(\mathbf{u}, \mathbf{v}) = \frac{\mathbf{u} \cdot \mathbf{v}}{\|\mathbf{u}\| \|\mathbf{v}\|}
\end{equation}

where $\mathbf{u} \cdot \mathbf{v}$ is the dot product of the two embeddings, and $\|\mathbf{u}\|$ and $\|\mathbf{v}\|$ are the magnitudes of the embeddings. A cosine similarity value closer to 1 indicates higher similarity, while a value closer to 0 indicates greater difference.

\paragraph{Accuracy}  
Accuracy measures the proportion of correctly retrieved items out of the total number of items. In retrieval tasks, it is often used to evaluate how often the correct documents are retrieved among all retrieved documents.

\[
\text{Accuracy}@k = \frac{\text{\# of Correctly Retrieved Items}}{k}
\]

\paragraph{Precision}  
Precision evaluates the proportion of relevant documents among the retrieved documents. It is a well-established and widely used metric in information retrieval and recommendation systems, because it effectively represents the scenarios where users are interested in the first few results returned by a system\cite{manning2008evaluation}. It is defined as:

\[
\text{Precision}@k = \frac{\text{\# of Relevant Documents Retrieved in Top k}}{k}
\]

\paragraph{Recall}  
Recall measures the proportion of relevant documents that were retrieved out of all relevant documents available. It is another important metric used in information retrieval, particularly when evaluating the completeness of the results returned by a retrieval system \cite{manning2008evaluation}\cite{izacard2022unsuperviseddenseinformationretrieval}\cite{xiong2020approximatenearestneighbornegative}. It is defined as:

\[
\text{Recall}@k = \frac{\text{\# Relevant Documents Retrieved in Top k}}{\text{Total Number of Relevant Documents}}
\]

\paragraph{NDCG (Normalized Discounted Cumulative Gain)}
NDCG evaluates the ranking quality of search results\cite{manning2008evaluation}. It accounts for the position of relevant documents in the result list, providing a single measure that balances the relevance of documents and their order, heavily penalizing relevant documents that appear lower in the results\cite{kekalainen2002using}. It is calculated as:

\[
\text{NDCG}@k = \frac{\text{DCG}@k}{\text{IDCG}@k}
\]

where DCG (Discounted Cumulative Gain) is:

\[
\text{DCG}@k = \sum_{i=1}^{k} \frac{2^{\text{rel}_i} - 1}{\log_2(i + 1)}
\]

In this equation, \( \text{rel}_i \) represents the relevance score of the document at position \( i \), and \( \log_2(i + 1) \) is used to discount the relevance score based on the position of the document.

IDCG (Ideal DCG) is the DCG score of the ideal ranking (i.e., the best possible ranking of documents):

\[
\text{IDCG}@k = \sum_{i=1}^{k} \frac{2^{\text{rel}^*_i} - 1}{\log_2(i + 1)}
\]

where \( \text{rel}^*_i \) is the relevance score of the document at position \( i \) in the ideal ranking.

\paragraph{MRR (Mean Reciprocal Rank)}
MRR measures the average rank of the first relevant document across multiple queries. It is used for evaluating the effectiveness of systems that return a ranked list of responses\cite{manning2008evaluation}\cite{kekalainen2002using}. It is defined as:

\[
\text{MRR}@k = \frac{1}{Q} \sum_{q=1}^{Q} \frac{1}{\text{rank}_q}
\]

where \( \text{rank}_q \) is the rank of the first relevant document for query \( q \) within the top \( k \) results, and \( Q \) is the total number of queries. If no relevant document is found in the top \( k \), the reciprocal rank is considered to be 0.

\paragraph{MAP (Mean Average Precision)}
MAP computes the average precision for each query and then averages these scores across all queries\cite{manning2008evaluation}. Average Precision (AP) for a single query is defined as:

\[
\text{AP}@k = \frac{1}{\min(R, k)} \sum_{i=1}^{k} (\text{Precision@i} \times \text{rel}_i)
\]

where \( R \) is the number of relevant documents for the query, \( \text{Precision}@i \) is the precision at rank \( i \), and \( \text{rel}_i \) is 1 if the document at rank \( i \) is relevant, otherwise 0.

MAP@k is the mean of the AP@k scores across all queries:

\[
\text{MAP}@k = \frac{1}{Q} \sum_{q=1}^{Q} \text{AP}@k_q
\]

where \( Q \) is the total number of queries.

\subsection{Target Models}
\paragraph{BERT} BERT (Bidirectional Encoder Representations from Transformers) employs a bidirectional approach, reading text in both directions simultaneously to capture deep contextual understanding \cite{reimers-2019-sentence-bert}.

\paragraph{Dense Passage Retrieval} Dense Passage Retrieval (DPR) uses a bi-encoder architecture to independently encode queries and passages into dense vectors, enabling semantic similarity-based retrieval \cite{karpukhin-etal-2020-dense}.

\paragraph{Contriever} Contriever generates high-quality embeddings for passages and queries by combining multiple training objectives and architectures to enhance retrieval accuracy \cite{izacard2022unsuperviseddenseinformationretrieval}.

\paragraph{SimCSE} SimCSE (Simple Contrastive Learning of Sentence Embeddings) fine-tunes language models using contrastive learning to improve sentence embeddings by optimizing the similarity between positive pairs and dissimilarity between negative pairs \cite{gao2022simcsesimplecontrastivelearning}.

\paragraph{ANCE} ANCE (Approximate Nearest Neighbor Negative Contrastive Learning) improves dense representations by using negative samples from an approximate nearest neighbor index, enhancing training efficiency and retrieval performance \cite{xiong2020approximatenearestneighbornegative}.

\subsection{Evaluation Methods}
\paragraph{Retrieval Effectiveness Evaluation} We use Sentence-BERT \cite{reimers2019sentencebertsentenceembeddingsusing} to generate the aforementioned metrics across selected datasets (NQ, Quora, FiQA,HotpotQA) from BEIR \cite{n2021beir}.

\paragraph{Retrieval Robustness Evaluation} We first try injecting adversarial passages into the corpus, using the method from \cite{zhong2023poisoningretrievalcorporainjecting}. However, the generation of a poisoned corpus requires massive computing resources. Thus, we quickly pivot to an innovative alternative: modifying the Sentence-Transformer from sentence-transformer python library\cite{reimers2019sentencebertsentenceembeddingsusing} by adding perturbation logic, which is used to simulate adversarial passages attack in queries. Specifically, we introduce random integer noise to the token IDs of tokenized text sequences input. Adversarial inputs are often used to evaluate the robustness of models\cite{kurakin2016adversarialexamplesinthephysicalworld}. This technique aims to test the robustness of each model by simulating errors, such as grammatical mistakes or typos, within the tokenized input. We then compare the result with the pre-perturbation result for each model to measure the drop in performance with this perturbation attack. We control the perturbation rate using a variable called \texttt{perturb\_rate}. If not specified, it defaults to 0.1, meaning 10\% of the input tokens are perturbed. To further evaluate robustness, we experiment with perturbation rates ranging from 5\% to 20\%, controlling the extent of the attack towards query input in retrieval systems. Furthermore, we have provided code to de-tokenize the perturbed input IDs to visualize how the perturbation affects the original text. Here is an example when the \texttt{perturb\_rate} is 0.10.

\begin{table}[h]
\begin{center}
\begin{tabularx}{\linewidth}{|X|X|}
\hline
\textbf{Previous Input} & \textbf{After Perturbation} \\
\hline\hline
[CLS] difference between a mcdouble and a double cheeseburger [SEP] [PAD] &  difference between a mcdou pay and a assembly sometimeburger [SEP] [PAD] \\
\hline
[CLS] what is theraderm used for [SEP] & [CLS] what is inraderm used for [SEP] \\
\hline
\end{tabularx}
\end{center}
\caption{De-tokenized Examples}
\label{tab:detokenized_examples}
\end{table}

\begin{algorithm}
\caption{Perturbation for Sentence-Transformer}
\label{alg:perturbation_logic}
\begin{algorithmic}[1]
\State \textbf{Input:} Sentences, Perturbation rate (\(\epsilon \in [0, 1]\))

\State \textbf{begin}
\For{each \texttt{query} in \texttt{encoded sentences}}
    \For{each \texttt{token} in \texttt{query}}
        \State \( r \gets \texttt{RandomUniform}(0, 1) \)
        \If{ \( r < \epsilon \) }
            \State \( d \gets \texttt{RandomInteger}(0, 9) \)
            \State \texttt{token} \(\gets\) \texttt{token} + \( d \)
        \EndIf
    \EndFor
\EndFor
\State \textbf{end}

\vspace{0.1cm}  % Adds vertical space

\State \textbf{Output:} Perturbed encoded input tokens from queries
\end{algorithmic}
\end{algorithm}
% \begin{algorithm}
% \caption{Perturbation for Sentence-Transformer}
% \label{alg:perturbation_logic}
% \begin{algorithmic}[1]
% \textbf{Input:} Sentences, Perturbation rate (\(\epsilon \in [0, 1]\))
% \hline
% \STATE \textbf{begin}
% \STATE \quad \textbf{for each} \texttt{query} \textbf{in} \texttt{encoded sentences} \textbf{do}
% \STATE \quad \quad \textbf{for each} \texttt{token} \textbf{in} \texttt{query} \textbf{do}
% \STATE \quad \quad \quad \( r \gets \texttt{RandomUniform}(0, 1) \)
% \STATE \quad \quad \quad \textbf{if} \( r < \epsilon \) \textbf{then}
% \STATE \quad \quad \quad \quad \( d \gets \texttt{RandomInteger}(0, 9) \)
% \STATE \quad \quad \quad \quad \texttt{token} \(\gets\) \texttt{token} + \( d \)
% \STATE \quad \quad \quad \textbf{end if}
% \STATE \quad \quad \textbf{end for}
% \STATE \quad \textbf{end for}
% \STATE \textbf{end}
% \vspace{0.1cm}  % Adds vertical space
% \hline
% \vspace{0.1cm}  % Adds vertical space
% \textbf{Output:} Perturbed encoded input tokens from queries
% \end{algorithmic}
% \end{algorithm}

\section{Experiments and Results}

\begin{table*}
\begin{center}
\begin{tabular}{lcccccc}
\hline
\textbf{Model} & \textbf{Acc@1} & \textbf{Prec@1} & \textbf{Rec@1} & \textbf{NDCG@10} & \textbf{MRR@10} & \textbf{MAP@100} \\
\hline
ANCE        & 0.77 & 0.77 & 0.50 & 0.73 & 0.81 & 0.68 \\
BERT        & 0.76 & 0.76 & 0.50 & 0.74 & 0.82 & 0.69 \\
Contriever  & 0.71 & 0.71 & 0.47 & 0.70 & 0.76 & 0.65 \\
DPR         & 0.52 & 0.52 & 0.35 & 0.51 & 0.57 & 0.47 \\
simCSE      & 0.34 & 0.34 & 0.25 & 0.35 & 0.39 & 0.32 \\
\hline
\end{tabular}
\end{center}
\caption{Avg. Performance (FiQA, HotpotQA, Quora)}
\label{tab:retrieval_performance}
\end{table*}

\begin{table*}
\begin{center}
\begin{tabular}{lcccccc}
\hline
\textbf{Model} & \textbf{Acc@1} & \textbf{Prec@1} & \textbf{Rec@1} & \textbf{NDCG@10} & \textbf{MRR@10} & \textbf{MAP@100} \\
\hline
ANCE-perturbed-5\%        & 0.029 & 0.029 & 0.018 & 0.027 & 0.027 & 0.028 \\
BERT-perturbed-5\%        & 0.042 & 0.042 & 0.025 & 0.039 & 0.040 & 0.039 \\
Contriever-perturbed-5\%  & 0.031 & 0.031 & 0.019 & 0.028 & 0.029 & 0.029 \\
DPR-perturbed-5\%         & 0.065 & 0.065 & 0.042 & 0.052 & 0.061 & 0.051 \\
simCSE-perturbed-5\%      & 0.031 & 0.031 & 0.021 & 0.028 & 0.032 & 0.025 \\
\hline
ANCE-perturbed-20\%       & 0.169 & 0.169 & 0.102 & 0.144 & 0.157 & 0.142 \\
BERT-perturbed-20\%       & 0.193 & 0.193 & 0.122 & 0.174 & 0.185 & 0.173 \\
Contriever-perturbed-20\% & 0.151 & 0.151 & 0.095 & 0.135 & 0.141 & 0.135 \\
DPR-perturbed-20\%        & 0.230 & 0.230 & 0.155 & 0.209 & 0.233 & 0.199 \\
simCSE-perturbed-20\%     & 0.135 & 0.135 & 0.096 & 0.129 & 0.148 & 0.119 \\
\hline
\end{tabular}
\end{center}
\caption{Avg. Performance Drop with 5\%, 20\% Perturbation (FiQA, HotpotQA, Quora)}
\label{tab:perturbed_retrieval_performance_drop}
\end{table*}

In our research, we used cosine similarity to evaluate the accuracy of various language models by measuring the closeness between the model's output and the semantic meaning in the validation dataset.

The results of our analysis revealed that ANCE and BERT exhibited superior performance in terms of accuracy. Both models demonstrated a high degree of alignment between their outputs and the semantic meaning of the input corpus. Contriever followed closely behind, with a slight drop in accuracy (as shown in Table \ref{tab:retrieval_performance}).

Furthermore, to assess the influence of perturbation attacks on the precision of language models, we subjected 5\% and 20\% of the input text to the attack. Our findings indicated that ANCE, Contriever, and simCSE were the least affected by this type of attack. Their accuracy remained relatively stable even after the input data was subjected to perturbation (Table \ref{tab:perturbed_retrieval_performance_drop}) .

However, it is worth noting that simCSE recorded a comparatively lower accuracy level compared to the other models. This observation suggests that simCSE may have encountered challenges in capturing the semantic meaning of the input corpus during the initial training phase. As a result, the perturbation attack had less of an impact on its performance since there was already limited accuracy to begin with.

For detailed output, please refer to Appendix section.

\section{Conclusion and Further Remarks}

Comparing the design of the mentioned retriever models, we find ANCE to be superior in effectiveness and robustness for several reasons:

\begin{itemize}
    \item \textbf{Dynamic Negative Sampling}: ANCE uses dynamic negative sampling, exposing the model to harder negatives during training. This enhances its ability to distinguish similar queries and documents, making it less susceptible to small perturbations and adversarial attacks. In contrast, BERT and Contriever lack this dynamic aspect.
    \item \textbf{Approximate Nearest Neighbor Search}: ANCE employs approximate nearest neighbor (ANN) search techniques (e.g., Faiss \cite{douze2024faiss}), allowing effective retrieval even with altered queries. This resilience to minor changes significantly contributes to its robustness against adversarial perturbations, unlike BERT and Contriever.
\end{itemize}

Understanding how models withstand adversarial attacks aids in making informed deployment choices, safeguarding the integrity and accuracy of LLM outputs. A future direction could be applying the above-mentioned designs to simple naive models and measuring the enhancement of robustness and reliability.

In conclusion, despite current limitations, we recognize the importance of robustness in retrieval models and aim to address this in future research to improve the safety and effectiveness of LLM applications.

\section{Appendix}
Table \ref{tab:fiqa_results}, Table \ref{tab:quora_results}, Table \ref{tab:quora_results}

% \begin{table*}
% \begin{center}
% \begin{tabular}{lcccccc}
% \hline
% \textbf{Model} & \textbf{Acc@1} & \textbf{Prec@1} & \textbf{Rec@1} & \textbf{NDCG@10} & \textbf{MRR@10} & \textbf{MAP@100} \\
% \hline
% ANCE        & 0.74 & 0.74 & 0.48 & 0.70 & 0.79 & 0.65 \\
% BERT        & 0.72 & 0.72 & 0.47 & 0.70 & 0.77 & 0.65 \\
% Contriever  & 0.68 & 0.68 & 0.45 & 0.67 & 0.74 & 0.62 \\
% DPR         & 0.45 & 0.45 & 0.31 & 0.46 & 0.51 & 0.41 \\
% simCSE      & 0.31 & 0.31 & 0.23 & 0.33 & 0.36 & 0.30 \\
% \hline
% \end{tabular}
% \end{center}
% \caption{Avg. Performance under 5\% Perturbation (FiQA, HotpotQA, Quora)}
% \label{tab:perturbed_retrieval_performance_5}
% \end{table*}

\begin{table*}
\begin{center}
\begin{tabular}{|l|c|p{8cm}|}
\hline
Student Name & Contributed Aspects & Details \\
\hline\hline
Ming Zhong & Implementation, Analysis, Paper Writing & 
Proposed and implemented evaluation pipeline for Contriever and BERT models; Analyzed results and extracted insights; Report and Proposal writing \\
Nanako Honda & Implementation and Analysis & Implemented simCSE model; Analyzed results and extracted insights; Report writing \\
Zhizhi Wu & Project Architecture, Implementation & Created project architecture and boilerplate code for calling inference API and IR evaluation. Implemented DPR and ANCE model E2E, implemented robustness evaluation and result visualization. \\
\hline
\end{tabular}
\end{center}
\caption{Contributions of team members.}
\label{tab:contributions}
\end{table*}

\section{Work Division}
Please refer to Table \ref{tab:contributions}.

{\small
\bibliographystyle{ieeetr}
\bibliography{egbib}
}

\begin{table*}
\begin{center}
\begin{tabular}{lcccccccc}
\hline
\textbf{Model} & \textbf{Acc@1} & \textbf{Prec@1} & \textbf{Rec@1} & \textbf{NDCG@10} & \textbf{MRR@10} & \textbf{MAP@100} \\ \hline
\textbf{ANCE} & 0.523 & 0.523 & 0.269 & 0.532 & 0.611 & 0.469 \\ 
\textbf{ANCE-perturbed-5\%} & 0.50 & 0.50 & 0.26 & 0.51 & 0.59 & 0.44 \\ 
\textbf{ANCE-perturbed-20\%} & 0.383 & 0.383 & 0.197 & 0.394 & 0.468 & 0.344 \\ \hline
\textbf{BERT} & 0.480 & 0.480 & 0.241 & 0.493 & 0.581 & 0.428 \\ 
\textbf{BERT-perturbed-5\%} & 0.44 & 0.44 & 0.23 & 0.46 & 0.54 & 0.40 \\ 
\textbf{BERT-perturbed-20\%} & 0.344 & 0.344 & 0.172 & 0.358 & 0.429 & 0.306 \\ \hline
\textbf{Contriever} & 0.316 & 0.316 & 0.145 & 0.343 & 0.417 & 0.284 \\ 
\textbf{Contriever-perturbed-5\%} & 0.30 & 0.30 & 0.14 & 0.33 & 0.40 & 0.27 \\ 
\textbf{Contriever-perturbed-20\%} & 0.252 & 0.252 & 0.116 & 0.268 & 0.334 & 0.221 \\ \hline
\textbf{DPR} & 0.142 & 0.142 & 0.070 & 0.157 & 0.194 & 0.132 \\ 
\textbf{DPR-perturbed-5\%} & 0.12 & 0.12 & 0.06 & 0.15 & 0.18 & 0.12 \\ 
\textbf{DPR-perturbed-20\%} & 0.088 & 0.088 & 0.042 & 0.100 & 0.129 & 0.080 \\ \hline
\textbf{simCSE} & 0.110 & 0.110 & 0.049 & 0.118 & 0.159 & 0.092 \\ 
\textbf{simCSE-perturbed-5\%} & 0.09 & 0.09 & 0.04 & 0.10 & 0.14 & 0.08 \\ 
\textbf{simCSE-perturbed-20\%} & 0.062 & 0.062 & 0.026 & 0.065 & 0.089 & 0.050 \\ \hline
\end{tabular}
\end{center}
\caption{Summary of Results for FIQA Dataset}
\label{tab:fiqa_results}
\end{table*}

\begin{table*}
\begin{center}
\begin{tabular}{lcccccccc}
\hline
\textbf{Model} & \textbf{Acc@1} & \textbf{Prec@1} & \textbf{Rec@1} & \textbf{NDCG@10} & \textbf{MRR@10} & \textbf{MAP@100} \\ \hline
\textbf{ANCE} & 0.952 & 0.952 & 0.820 & 0.969 & 0.969 & 0.959 \\ 
\textbf{ANCE-perturbed-5\%} & 0.93 & 0.93 & 0.80 & 0.95 & 0.95 & 0.94 \\ 
\textbf{ANCE-perturbed-20\%} & 0.810 & 0.810 & 0.700 & 0.854 & 0.848 & 0.834 \\ \hline
\textbf{BERT} & 0.946 & 0.946 & 0.816 & 0.964 & 0.965 & 0.952 \\ 
\textbf{BERT-perturbed-5\%} & 0.90 & 0.90 & 0.78 & 0.93 & 0.93 & 0.91 \\ 
\textbf{BERT-perturbed-20\%} & 0.733 & 0.733 & 0.632 & 0.780 & 0.774 & 0.758 \\ \hline
\textbf{Contriever} & 0.943 & 0.943 & 0.813 & 0.964 & 0.964 & 0.952 \\ 
\textbf{Contriever-perturbed-5\%} & 0.91 & 0.91 & 0.79 & 0.94 & 0.94 & 0.92 \\ 
\textbf{Contriever-perturbed-20\%} & 0.771 & 0.771 & 0.666 & 0.826 & 0.818 & 0.803 \\ \hline
\textbf{DPR} & 0.769 & 0.769 & 0.660 & 0.803 & 0.811 & 0.776 \\ 
\textbf{DPR-perturbed-5\%} & 0.69 & 0.69 & 0.59 & 0.73 & 0.73 & 0.70 \\ 
\textbf{DPR-perturbed-20\%} & 0.436 & 0.436 & 0.373 & 0.488 & 0.486 & 0.462 \\ \hline
\textbf{simCSE} & 0.725 & 0.725 & 0.620 & 0.772 & 0.777 & 0.741 \\ 
\textbf{simCSE-perturbed-5\%} & 0.68 & 0.68 & 0.58 & 0.73 & 0.74 & 0.70 \\ 
\textbf{simCSE-perturbed-20\%} & 0.475 & 0.475 & 0.407 & 0.535 & 0.534 & 0.505 \\ \hline
\end{tabular}
\end{center}
\caption{Summary of Results for Quora Dataset}
\label{tab:quora_results}
\end{table*}

\begin{table*}
\begin{center}
\begin{tabular}{lcccccccc}
\hline
\textbf{Model} & \textbf{Acc@1} & \textbf{Prec@1} & \textbf{Rec@1} & \textbf{NDCG@10} & \textbf{MRR@10} & \textbf{MAP@100} \\ \hline
\textbf{ANCE} & 0.821 & 0.821 & 0.411 & 0.682 & 0.862 & 0.606 \\ 
\textbf{ANCE-perturbed-5\%} & 0.78 & 0.78 & 0.39 & 0.65 & 0.82 & 0.57 \\ 
\textbf{ANCE-perturbed-20\%} & 0.596 & 0.596 & 0.298 & 0.503 & 0.657 & 0.430 \\ \hline
\textbf{BERT} & 0.867 & 0.867 & 0.434 & 0.756 & 0.899 & 0.691 \\ 
\textbf{BERT-perturbed-5\%} & 0.82 & 0.82 & 0.41 & 0.71 & 0.86 & 0.65 \\ 
\textbf{BERT-perturbed-20\%} & 0.638 & 0.638 & 0.319 & 0.553 & 0.688 & 0.488 \\ \hline
\textbf{Contriever} & 0.876 & 0.876 & 0.438 & 0.782 & 0.913 & 0.714 \\ 
\textbf{Contriever-perturbed-5\%} & 0.83 & 0.83 & 0.42 & 0.74 & 0.88 & 0.67 \\ 
\textbf{Contriever-perturbed-20\%} & 0.659 & 0.659 & 0.330 & 0.591 & 0.719 & 0.521 \\ \hline
\textbf{DPR} & 0.636 & 0.636 & 0.318 & 0.562 & 0.702 & 0.488 \\ 
\textbf{DPR-perturbed-5\%} & 0.54 & 0.54 & 0.27 & 0.49 & 0.61 & 0.42 \\ 
\textbf{DPR-perturbed-20\%} & 0.333 & 0.333 & 0.167 & 0.309 & 0.394 & 0.259 \\ \hline
\textbf{simCSE} & 0.176 & 0.176 & 0.088 & 0.173 & 0.228 & 0.138 \\ 
\textbf{simCSE-perturbed-5\%} & 0.15 & 0.15 & 0.07 & 0.15 & 0.19 & 0.12 \\ 
\textbf{simCSE-perturbed-20\%} & 0.069 & 0.069 & 0.035 & 0.075 & 0.097 & 0.058 \\ \hline
\end{tabular}
\end{center}
\caption{Summary of Results for HotpotQA Dataset}
\label{tab:hotpotqa_results}
\end{table*}

\end{document}